\pdfoutput=1
\documentclass[11pt]{article}

\usepackage[]{acl}

\usepackage{times}
\usepackage{latexsym}

\usepackage[T1]{fontenc}
\usepackage[utf8]{inputenc}

\usepackage{microtype}

\usepackage{inconsolata}

\usepackage{algorithm}
\usepackage{algpseudocode}

\usepackage{algorithmicx}
\usepackage{amsmath}
\usepackage{amsfonts}
\usepackage{amssymb}
\usepackage{enumitem}
\usepackage{multicol}
\usepackage{multirow}
\usepackage{array}
\usepackage{booktabs}
\usepackage{float}
\usepackage{setspace}
\usepackage{tabularx}
\usepackage{graphicx}
\usepackage{amssymb} % 提供 \checkmark 命令
\usepackage{pifont}   % 提供 \ding 命令
\usepackage{xcolor}   % 提供颜色选项

\usepackage{amsmath,amsfonts,bm}

\def\eqref#1{equation~\ref{#1}}
\def\1{\bm{1}}

\DeclareMathAlphabet{\mathsfit}{\encodingdefault}{\sfdefault}{m}{sl}
\SetMathAlphabet{\mathsfit}{bold}{\encodingdefault}{\sfdefault}{bx}{n}

\newcommand\ka{$\textsc{ODA}$}
\newcommand\kas{$\textsc{ODA}$\ }

\title{ODA: Observation-Driven Agent for integrating LLMs and Knowledge Graphs}

\author{
  Lei Sun\footnotemark[1]\footnotemark[2]\textsuperscript{1}\quad Zhengwei Tao\footnotemark[1] \textsuperscript{2}\quad Youdi Li \textsuperscript{1}\quad Hiroshi Arakawa \textsuperscript{1}\\
  \textsuperscript{1}Panasonic Connect Co., Ltd., Japan \\
  \textsuperscript{2}Peking University \\
  \texttt{tttzw@stu.pku.edu.cn} \\
  \texttt{\{sun.lei, ri.yutei, arakawa.hrs\}@jp.panasonic.com} 
}

\begin{document}
\maketitle
\thispagestyle{plain} % 在标题页添加页码
\begin{abstract}
The integration of Large Language Models (LLMs) and knowledge graphs (KGs) has achieved remarkable success in various natural language processing tasks. However, existing methodologies that integrate LLMs and KGs often navigate the task-solving process solely based on the LLM's analysis of the question, overlooking the rich cognitive potential inherent in the vast knowledge encapsulated in KGs. To address this, we introduce Observation-Driven Agent (\ka), a novel AI agent framework tailored for tasks involving KGs. \kas incorporates KG reasoning abilities via global observation,  which enhances reasoning capabilities through a cyclical paradigm of observation, action, and reflection. Confronting the exponential explosion of knowledge during observation, we innovatively design a recursive observation mechanism. Subsequently, we integrate the observed knowledge into the action and reflection modules. Through extensive experiments, \kas demonstrates state-of-the-art performance on several datasets, notably achieving accuracy improvements of 12.87\% and 8.9\%. Our code and data are available on \href{https://github.com/lanjiuqing64/KGdata}{ODA}.

% Coupling Large Language Models  (LLM) and knowledge graphs (KG) has achieved remarkable success in various natural language processing tasks. However, existing methodologies integrating the LLM and KG navigate the task-solving process by merely relying on the LLM's analysis of the question, overlooking the rich cognitive potential inherent in the abundant knowledge encapsulated in KGs. To address this, we introduce Observation-Driven Agent (\ka), a novel AI agent framework tailored for tasks of knowledge graphs. \kas incorporates KG reasoning abilities via global observation that enhances reasoning capabilities through a cyclical paradigm of observation, action, and reflection. Confronting the exponential explosion of observation knowledge, we innovatively design a recursive observation mechanism. Subsequently, we integrate the observed knowledge into the action and reflection modules.Through extensive experiments, \kas demonstrates state-of-the-art performance on several datasets, notably achieving accuracy improvements of 12.87\% and 8.9\%. 
\end{abstract}

\renewcommand{\thefootnote}{\fnsymbol{footnote}}
\footnotetext[1]{Equal contribution}
\footnotetext[2]{Corresponding author}
\section{Introduction}

Large language models (LLMs)~\cite{touvron2023llama, scao2022bloom, muennighoff2022crosslingual, brown2020language} have exhibited extraordinary capabilities across a variety of natural language processing tasks. 
Despite their impressive accomplishments, LLMs often struggle to provide accurate responses to queries that necessitate specialized expertise beyond their pre-training content. 
In response to this limitation, a natural and promising approach involves the integration of external knowledge sources, such as knowledge graphs (KGs), to augment LLM reasoning abilities. KGs provide structured, explicit, and explainable knowledge representations, offering a synergistic method to overcome the intrinsic constraints of LLMs. The fusion of LLMs with KGs has garnered significant interest in recent research~\cite{pan2024unifying}, underlying a vast array of applications~\cite{zhang2023making, do2024constraintchecker, sun2023think}.

\begin{figure}
    \centering
    \includegraphics[width=1\linewidth]{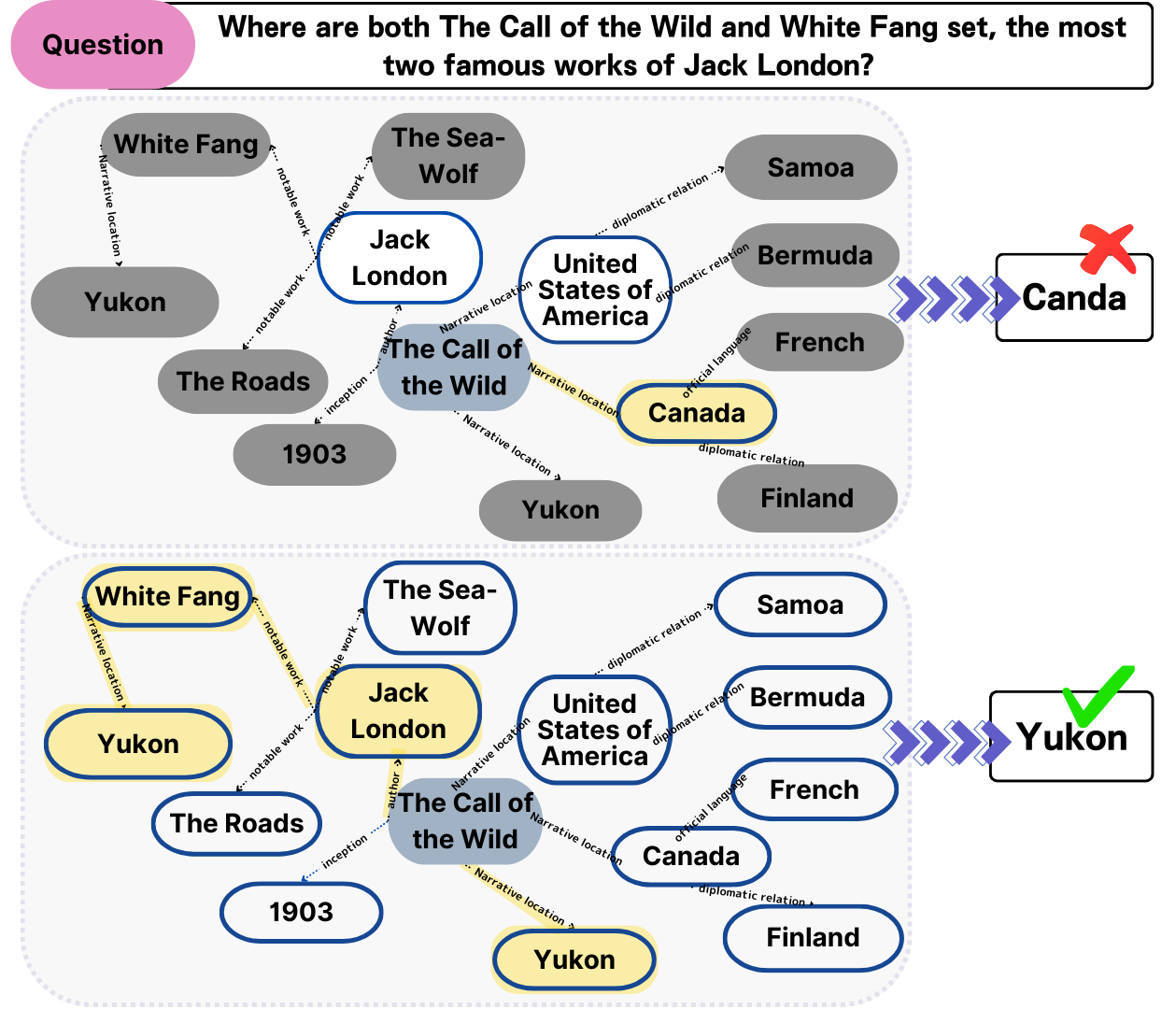}
    \caption{An example of LLM integrating with KG. Observed entities are shown in white, while non-observed entities are displayed in gray. Entities selected by the agent to answer the question are highlighted in yellow.}
    \label{fig:intro-pic}
\end{figure}

Existing methodologies for solving tasks that integrate KGs with LLMs can be categorized into two groups. The first one involves retrieving relevant triples from KGs in response to specific questions~\cite{wang2023boosting,luo2024reasoning, jiang2023structgpt}. The second part adopts an explore-exploit strategy, directing the knowledge utilization process within the graph according to the question~\cite{sun2023think, guo2023knowledgenavigator}. However, both categories navigate the task-solving process by merely relying on the LLM's analysis of the question, overlooking the rich cognitive potential inherent in the abundant knowledge encapsulated in KGs. 
KGs, which store a wealth of informative and symbolic facts, should deeply participate in the reasoning process together with LLM rather than being merely treated as a static repository of knowledge~\cite{pan2024unifying}. 
As the example in the upper panel of Figure~\ref{fig:intro-pic}, LLM analyzes the question and navigates towards \textit{Narrative location} relation of entity \textit{The Call of The Wild}. However, this entity has many neighboring entities with that relation, leading LLM to incorrectly infer \textit{Canada} as the answer. In contrast, the bottom panel demonstrates how KG provides key patterns that reveal both \textit{The Call of The Wild} and \textit{White Fang} share the location \textit{Yukon}.
If LLM could observe this information beforehand, it would precisely guide its reasoning process towards the correct answer (as shown in the bottom panel). 
Therefore, LLM should adopt an overall observation to incorporate the extensive knowledge and intricate patterns embedded within the KG. Achieving this objective presents two primary challenges: firstly, a global observation of the KG can result in an exponential growth in the number of triples. As shown in the upper panel of Figure \ref{fig:intro-pic}, fully processing all 3-hop connections for \textit{The Call of the Wild} is impractical. Secondly, the integration of such comprehensive observation into the existing reasoning paradigms of LLMs presents another challenge. How to combine the observation with the reasoning process of LLM matters for solving the tasks.
% In the bottom panel of Figure \ref{fig:intro-pic}, although we can leverage the complete knowledge within a 3-hop radius of \textit{The Call of the Wild}, the LLM's reasoning process may suffer from insufficient integration of observational information. 
% This can hinder its ability to identify the crucial paths, particularly the highlighted yellow nodes, which play a vital role in determining the correct answer \textit{Yukon}.  

% 用图中例子解释

% To mitigate these deficients, we propose a novel integration of LLM and KG namely \ka, an AI Agent tailored for KG processing. \ka solves tasks with KG in an observe, act, and reflection cycleling paradigm. To incorporate the reasoning capability of KG, \ka first explore the KG according to the current state. We design a observation method to efficiently draw patterns and avoid the problem of exponential growth of nodes. 
% After the observation, \ka take action with fusion of LLM thoughts and the observed KG patterns. 
% Then \ka reflect its internal states with the action results and the observation. This process iterates util \ka is able to complete the task.

% Motivated by that, we introduce a novel framework for the sufficient integration abilities of both LLM and KG, namely Observation-Driven Agent~(\ka), an AI Agent specifically designed for KG-centric tasks. 
Motivated by this, we introduce a novel framework, the Observation-Driven Agent (ODA), aimed at sufficiently and autonomously integrating the capabilities of both LLM and KG. ODA serves as an AI Agent specifically designed for KG-centric tasks.
\kas engages in a cyclical paradigm of observation, action, and reflection. Within \ka, we design a novel observation module to efficiently draw autonomous reasoning patterns of KG. 
Our observation module avoids the problem of exponential growth of triples via recursive progress. 
% It finally extracts a concise and accurate subgraph from the vast KG. 
This approach ensures \kas integrating abilities of KG and LLM while mitigating the challenges associated with excessive data in KG, improving the efficiency and accuracy.
% 补充
Following the observation phase, \kas takes action by autonomously amalgamating insights derived from LLM inferences with the observed KG patterns. \kas can perform actions of three distinct types: Neighbor Exploration, Path Discovery, and Answering.
% 补充
Subsequently, \kas reflects on its internal state, considering both the outcomes of its actions and the prior observations. This iterative process continues until \kas accomplishes the task at hand.

We conduct extensive experiments to testify to the effectiveness of \kas on four datasets: \texttt{QALD10-en}, \texttt{T-REx}, \texttt{Zero-Shot RE} and \texttt{Creak}. Notably, our approach achieved state-of-the-art (SOTA) performance compared to competitive baselines. Specifically, on \texttt{QALD10-en} and \texttt{T-REx} datasets, we observed remarkable accuracy improvements of 12.87\% and 8.9\%, respectively. We conclude the contributions as follows:
\begin{itemize}[topsep=0pt]
\setlength{\parskip}{0pt}
\setlength{\parsep}{-1pt}
\setlength{\leftmargin}{-1pt}
\item[$\bullet$] We propose \ka, an AI Agent tailored for KG-centric tasks. \kas conducts observation to incorporate the reasoning ability of KG.

\item[$\bullet$] We design action and reflection modules that integrate observation into LLM reasoning. This strategy leverages the autonomous reasoning of KG and LLM in synergy.
% \item[$\bullet$] We further design action and reflection modules to integrate observation into LLM reasoning to exploit the capabilities of both sides.

\item[$\bullet$] We conduct experiments on four datasets and achieve SOTA performances.

\end{itemize}

\begin{figure*}[t]
    \centering
    \includegraphics[width=1\linewidth]{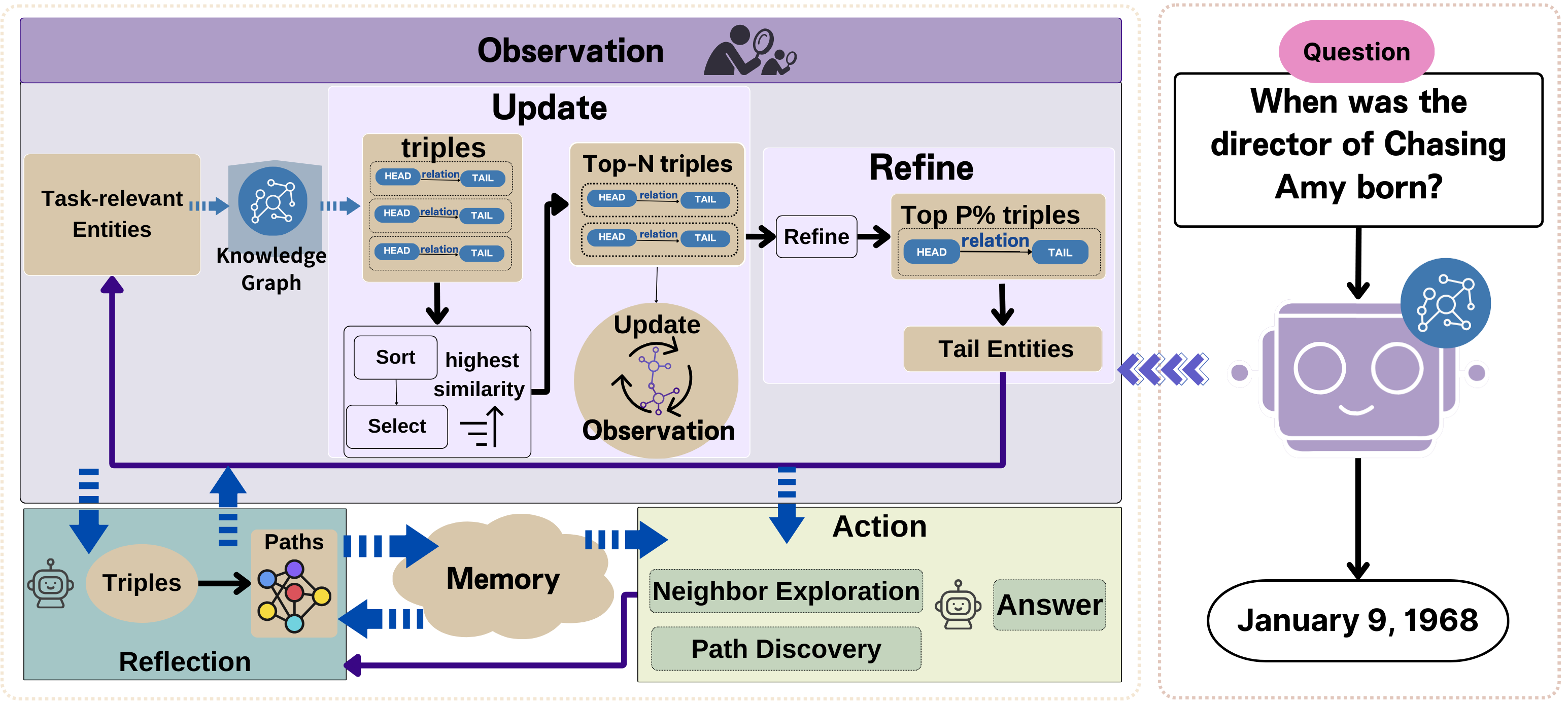}
    \caption{The overall framework of \ka.}
    \label{fig:workflow}
\end{figure*}

\section{Methods}
In this work, we aim to solve tasks associated with KG.
Let \( q \) represent a user question. The task \( T \) can be defined as generating an answer $Y$ given a question $q$, task-relevant entities $E=\{e_0,e_1,...,e_k\}$, and a KG denoted as \( G \). Formally, the task \( T \) can be expressed as: 
\[ T: (q,E), G \rightarrow Y \]

\begin{figure*}[t]
    \centering
    \includegraphics[width=1\linewidth]{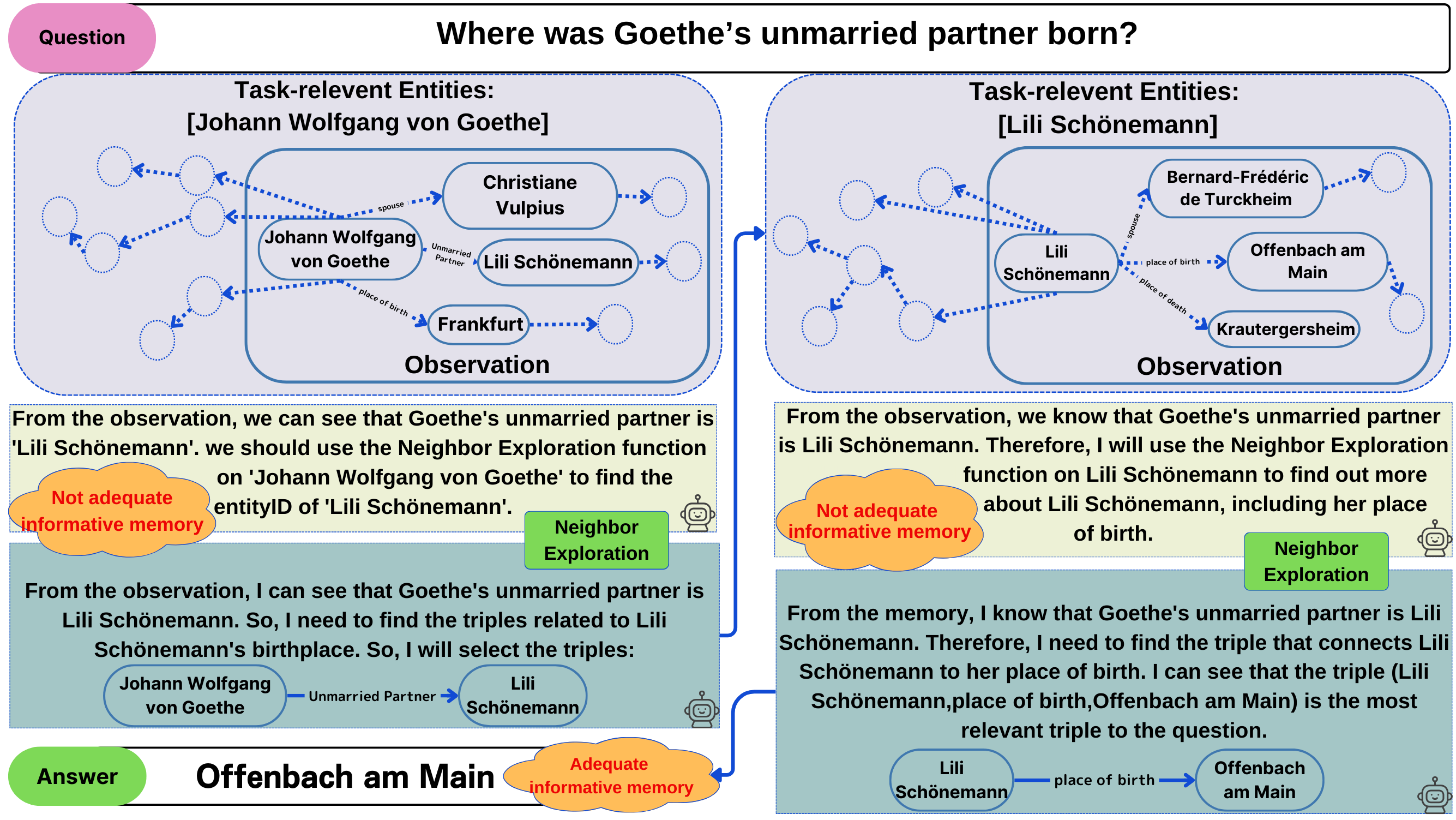}
    \caption{An example workflow of \ka. In this case, \kas initiates the obervation with entity \textit{ Johann Wolfgang von Goethe}. During the first iteration on the left side, the Neighbor Exploration of \textit{ Johann Wolfgang von Goethe} is selected, and the reflected triple \textit{(Johann Wolfgang von Goethe, unmarried Partner, Lili Schöneman)} is stored in memory. Subsequently, The observation of \textit{Lili Schöneman} then guides \kas to choose Neighbor Exploration action, and leads to the retention of the triple \textit{(Lili Schöneman, place of birth, Offenbach am Main)} in memory, as shown on the right side. Once sufficient knowledge has been accumulated, \kas triggers the Answer action, correctly identifying \textit{Offenbach am Main} as the answer.}
    \label{fig:caseworkflow}
\end{figure*}
% \vspace{-10pt}

Employing an iterative approach, \kas tackles the challenges inherent in KG-centric tasks. In contrast to existing methods that couple LLMs and KGs and rely solely on analyzing the LLM's query, \kas autonomously integrates observed knowledge from the KG into the entire reasoning process, resulting in more informed decisions. To achieve this objective, our \kas system, illustrated in Figure \ref{fig:workflow}, primarily comprises three key modules for task resolution: 

\begin{itemize}
    \item \textbf{Observation}: This module efficiently observes and processes relevant knowledge from the KG environment. In each iteration $i$, it constructs an observation subgraph (denoted as $O_i$). By leveraging insights and patterns gleaned from the KG, this subgraph is autonomously incorporated into a reasoning LLM. This synergistic integration equips \kas with enhanced capabilities from both the LLM and KG, allowing it to tackle tasks more effectively.
    \item \textbf{Action}: Drawing upon both the observation subgraph $O_i$ and \kas memory (denoted as $M_{<i}$), the action module,represented by $a_i$, strategically selects the most suitable action to execute on the KG, ensuring the accurate answering of the question.
    \item \textbf{Reflection}: Utilizing the observation subgraph $O_i$, the reflection module provides feedback by reflecting on the knowledge obtained from the action step. The reflected knowledge is then stored in memory $M_i$ for the next iteration, facilitating continuous reasoning.
\end{itemize}
Through this iterative process, \kas dynamically updates its observation subgraph $O_i$ and memory $M_i$ at each iteration \(i\). Each module is discussed in detail in the following sections.

\subsection{Observation}
The observation module is designed to inspect global KG knowledge and navigate the autonomous reasoning process with the KG environments. At each iteration \(i\), it leverages task-relevant entities \(E_i\) and a question \(q\) to generate an observation subgraph \(O_i\). This process can be formulated as:
\[O_i = \text{Observation}([E_i, q])\]
Initially, the task-relevant entities are populated with the entities embedded within the question \(q\).

For KG-centric tasks, the observation incurs the problem of an explosive number of nodes. To address the scalability challenge during observation subgraph updates, we propose a $D$-turn observe strategy, where $D$ represents the maximum hop depth. Each turn has two steps: update and refine. The update step focuses on expanding the subgraph, while the refining step ensures its appropriate size without loss of important information.

For each entity $e\in E_i$, the observation module initializes the observation entities as a set \(E_o^d = \{e\}\) at hop depth \(d\), where \(d\) represents the current search depth within the KG. The two-step, update and refine, iterates for each entity \(e\) until \(D\) is reached. The specific details are described as:

\begin{algorithm}[htbp]
\caption*{\textbf{Algorithm 1} Observation}
% \begin{itemize}
\textbf{Require:} Question $q$, limit $D$, $N$, and $P$
% \end{itemize}
\begin{algorithmic}[ ]
\State Initialize task-relevant entities \(E_i\) with the entities in $q$ 
\For{$e \in E_{i}$}
    \State Set $d = 0$
    \State Initialize observation entities ${E_o^d} = \{e\}$
    \While{$d < D$}
        \For{$ entity \in {E_o^d}$}
            \State Extract the neighboring triples
            \EndFor
        \For{$(r,t) \in \text{triples}$}
            \State Cosine similarity($q$,$r+t$)
        \EndFor
        \State Sort similarity scores of triples
        \State Append top ${N}$ triples to $O_i$
        \State Extract top ${P\%}$ triples from top ${N}$
        \State Update ${E_o^d}$ with  $t$ in top ${P\%}$
        \State Increment $d$
    \EndWhile
\EndFor
\end{algorithmic}
\end{algorithm}
% \vspace{-10pt}

\textbf{Update}:
\begin{itemize}
    \item For each entity $e$ in \(E_o^d\), neighboring triples are extracted from KG. Each triple takes the form $[e,r,t]$, where $r$ signifies the relation, and $t$ denotes the tail entity. 
    \item The similarity score between the question and the combined representation of $r$ and $t$, is computed by measuring the cosine similarity of their embeddings\footnote[1]{We use the GPT text-embedding-ada-002 model from OpenAI for encoding.}:
    \vspace{-1pt}
    \[
    \text{Cosine Similarity}(\mathbf{v}_q, \mathbf{v}_{r+t}) = \frac{\mathbf{v}_q \cdot \mathbf{v}_{r+t}}{\|\mathbf{v}_q\| \|\mathbf{v}_{r+t}\|}
    \]
    \item All triples associated with entities in $E$ are collectively sorted in descending order based on their similarity scores.
    \item The Top-$N$ triples are added to the observation subgraph \(O_i\).
\end{itemize}

\textbf{Refine}:
\begin{itemize}
    \item The Top-$N$ triples are further refined by retaining only the top \(P\%\) with the highest similarity scores.
    \item The tail entities from the refined top \(P\%\) triples are identified as the starting observation entities \(E_o^d\) for the next iteration.
\end{itemize}

% Then we obtain the observation subgraph \(O_i\) for $i^{th}$ turn. It is then utilized for the next modules.

% \subsection{Memory}

% Memory serves as a vital long-term storage component essential for KG-related tasks. This memory $M$, guides both the subsequent action and reflection modules. In interaction with the action module, memory $M$ aids in selecting the most suitable action to ensure successful task completion. 

\subsection{Action}

% Within an AI Agent framework, memory plays a crucial role as a long-term storage component \cite{}. In our \kas, memory $M$ is involved in guiding both the action and reflection modules. In its interaction with the action module, memory assists in selecting the most appropriate action, thereby facilitating successful task completion.
% Within an AI Agent framework, memory plays a crucial role as a long-term storage component \cite{shinn2023reflexion, yao2023retroformer}. In our \kas, memory $M_{<i}$ is involved in guiding both the action and reflection modules. In its interaction with the action module, memory assists in selecting the most appropriate action, thereby facilitating successful task completion.

Harnessing the power of an LLM, the action module crafts strategic prompts to generate optimal actions. Based on its memory \(M_{<i}\), observation subgraph \(O_i\), and historical actions \(a_{<i}\), the \kas selects the most accurate action \(a_i\).
\[a_i = \text{Action}([O_i, a_{<i}, M_{<i}])\]

We propose three core actions designed to empower \kas:

\begin{itemize}
    \item \textbf{Neighbor Exploration}: This action explores the KG neighborhood of task-relevant entities \(E_{i}\) and retrieves all neighboring triples. This helps build context and understand interconnectedness within the KG for \kas. 
    \item \textbf{Path Discovery}: Given two entities in task-relevant entities \(E_{i}\), this action searches for all possible paths connecting them. Each path consists of interconnected triples, allowing the \kas to explore various connections and potentially uncover hidden relationships.
    \item \textbf{Answer}: This action responds to the question only if the required information is present in memory $M_{<i}$.
\end{itemize}

Upon selecting an answer action, \kas halts the iterative loop of observation, action, and reflection. Leveraging the reliable knowledge within memory \(M_{<i}\), it can then directly formulate the answer to the question.
Alternatively, if a Neighbor Exploration or Path Discovery action is selected, \kas strategically extracts relevant knowledge from the KG as a set of triples. These extracted triples are then fed into the subsequent reflection step for further processing. The prompt used here can be found in Table \ref{tab:actionprompt}.

\subsection{Reflection}
% Furthermore, the reflection module relies on memory $M$ to accurately evaluate the output of actions. Designed specifically for KG tasks, memory M adopts a format comprising a network of paths that align with the innate structure of KGs, aimed at optimizing efficiency and relevance. Each path within $M$ encompasses a set of triples formatted as $[e,r,t]$, where $t$ signifies the relation, and $r$ denotes the tail entity.

The reflection module plays a crucial role in evaluating the triples generated from the action step and subsequently updating \kas memory \(M_i\). Designed specifically for KG tasks, memory \(M_i\) adopts a subgraph format consisting of a network of paths that align with the inherent structure of KG, aimed at optimizing efficiency and relevance. By integrating the observation subgraph \(O_i\) and existing memory \(M_{<i}\) autonomously, the reflection module provides invaluable feedback that guides future decision-making. This process can be formalized as:
\[ M_i = \text{Reflection}([O_i, a_i, M_{<i}]) \]

Given that memory \(M_i\) is structured as a network of paths, the reflection module navigates these paths to identify the first suitable one for integrating the reflected triple. This suitability arises from aligning the tail $t$ of the last triple in the selected path with the entity $e$ of the reflected triple. If a matching path is found, the reflected triple is appended. Otherwise, a new path is created based on the reflected triple. The maximum size of reflected triples is denoted as $K$. 

Subsequently, the tail entities in the reflected triples are designated as the task-relevant entities for the next iteration. The specific prompt description used for the reflection module is provided in Table \ref{tab:reflprompt}.

The observation, action, and reflection modules collaborate iteratively until either the Answer action is triggered or the maximum iteration limit is reached. Figure \ref{fig:caseworkflow} shows how observation, action, and reflection work together.

\section{Experiments}
\begin{table}[htbp]
\small
\setlength\tabcolsep{2pt}
\begin{tabular}{lcccc}
\toprule
Dataset & Test & Entity & Type & License \\
\midrule
\texttt{QALD10-en}         & 333           & 396             & Multi-hop      & MIT License        \\ 
\texttt{T-REx}             & 5000          & 4943            & Slot-Filling   & MIT License   \\ 
\texttt{Zero-Shot RE}      & 3724          & 3657            & Slot Filling   & MIT License          \\ 
\texttt{Creak}             & 1371          & 516             & Fact Checking  & MIT License           \\ 
\bottomrule
\end{tabular}
\caption{Dataset statistics. \textbf{Entity} stands for the entity size derived from all the question within the datasets.}
\label{tab:dataset}
\end{table}
\vspace{-10pt}

\subsection{Dataset}
To evaluate the performance of our \kas, we conduct experiments on four KBQA datasets: \texttt{QALD10-en}  \cite{perevalov2022qald9plus},\texttt{Creak}  \cite{onoe2021creak}, \texttt{T-REx} \cite{elsahar-etal-2018-rex}, and \texttt{Zero-Shot RE}  \cite{petroni-etal-2021-kilt}. Detailed specifications for each dataset are provided in Table \ref{tab:dataset}. The Hits@1\cite{sun2019rotate} accuracy with exact match is utilized as our evaluation metric.

\subsection{Setup}

We utilized the GPT-4 \cite{openai2023gpt} model as the \kas via the OpenAI API. Throughout our experiments, we consistently configured the temperature value of GPT-4 to 0.4 and set the maximum token length to 500.

For the observation step, we tuned key parameters based on preliminary experiments. we set $P_t$ to 10 and $N_t$ to 50. Furthermore, the \kas loop was capped at a maximum of 8 iterations. Lastly, the maximum hop depth $D$ is set to 3. As for the reflection module, we set the size of reflected triples $K$ to 15.

To establish Wikidata KG database and retrieve information from it, we employed the \textit{simple-wikidata-db}\footnote[1]{https://github.com/neelguha/simple-wikidata-db} Python library. This library provides various scripts for downloading the Wikidata dump, organizing it into staging files, and executing distributed queries on the data within these staged files. Specifically, we deployed the Wikidata dump across five AWS EC2 instances, each consisting of a 768GB machine with 48 cores.

Considering that our \kas relies heavily on continuous interaction with the KG, we discovered that the real-time extraction of required Wikidata knowledge on AWS achieved an average completion time of 50 seconds per question-answer pair within \texttt{QALD10-en} dataset. However, as the KBQA dataset expanded, the cost of using the Wikidata database on AWS became prohibitively expensive. Consequently, to address the computational expenses involved, we devised a solution by generating an offline subgraph for each KBQA dataset. This offline subgraph captures all the triples within a 3-hop radius of the entities in each dataset, including the properties of both the entities and the relations involved. Notably, generating such a subgraph for the \texttt{T-REx}  dataset, with its 4943 entities (as listed in Table \ref{tab:dataset}), takes approximately 54 minutes and 42.834 seconds in practice.

% The subgraph for each KBQA dataset is publicly available at \href{https://github.com/lanjiuqing64/KGdata}{https://github.com/lanjiuqing64/KGdata}.

\subsection{Baseline Models}

To comprehensively evaluate \kas effectiveness, we conduct a rigorous benchmark against several SOTA models across diverse categories. The comparison encompasses various models, starting with prompt-based approaches that do not utilize external knowledge. These include direct answering with GPT-3.5 and GPT-4, as well as the Self-Consistency \cite{wang2023selfconsistency} and CoT \cite{sun2023think}. On the other hand, Kownledge-combined models are considered, which incorporate fine-tuned techniques such as SPARQL-QA \cite{borroto2022sparqlqa}, RACo \cite{yu-etal-2022-retrieval}, RAG \cite{petroni-etal-2021-kilt} and Re2G \cite{glass-etal-2022-re2g}. Additionally, there is ToG \cite{sun2023thinkongraph} model, which integrates LLM with KG to bolster question-answering proficiency.

\subsection{Main Result}

Our \kas method outperforms existing methods, as shown in Table \ref{tab:main}. On average, our method achieves an accuracy gain of up to 19.58\% compared to direct answering with GPT-4, 19.28\% compared to fine-tuned models, and 7.09\% compared to TOG. These results demonstrate the efficiency and effectiveness of our method in comparison to other state-of-the-art methods.

Furthermore, our \kas significantly outperforms the prompt-based methods across various datasets, particularly showing an improvement of 65.50\% and 23.77\% on \texttt{Zero-Shot REx}  and \texttt{QALD10-en} , respectively. These results underscore the importance of leveraging external knowledge graphs for reasoning and completing the question-answering task.

Compared to the fine-tuned method, our \kas method demonstrates superior performance. Specifically, our method achieves a performance gain of 21.27\% for the \texttt{QALD10-en}  dataset, 6.99\% for the \texttt{Creak}  dataset, and 50.56\% for the \texttt{Zero-Shot RE}  dataset. Notably, this interaction between the LLM and KG, as our method employs, proves more effective than data-driven fine-tuned techniques, despite requiring no explicit training.

Our \kas method exhibits significant performance gains over the ToG method across most datasets, with improvements of 12.87\% (\texttt{QALD10-en}), 8.9\% (\texttt{T-REx}), and 7\% (\texttt{Zero-Shot RE}), despite both methods leveraging large language models and knowledge graphs. This performance disparity highlights the critical role of our observation module and the effectiveness of autonomously incorporating reasoning from KG. Specifically, our method demonstrates significantly stronger performance on the \texttt{QALD10-en}  dataset, known for its multi-hop and complex reasoning requirements. This achievement underscores our ODA ability to exploit the rich knowledge and patterns within KG effectively, combining the autonomous reasoning strengths of both LLM and KG to tackle complex questions successfully.
\begin{table*}[ht]
\centering
\small
% \setstretch{1} % 调整行间距
\begin{tabularx}{\textwidth}{l*{7}{>{\centering\arraybackslash}X}}
\toprule
Method & \texttt{QALD10-en} & \texttt{Creak} & \texttt{T-REx} & \texttt{Zero-Shot RE} & Average \\
\midrule
\multicolumn{6}{c}{w.o. Knowledge} \\
\midrule
Direct answering(GPT3.5) & \small 44.74\textsuperscript{\phantom{(1)}} & \small 90.00\textsuperscript{\phantom{(1)}} & \small 37.78\textsuperscript{\phantom{(1)}} & \small 37.14\textsuperscript{\phantom{(1)}} & \small 52.42\\
Direct answering(GPT4) & \small 57.10\textsuperscript{\phantom{(1)}} & \small 94.52\textsuperscript{\phantom{(1)}} & \small 57.72\textsuperscript{\phantom{(1)}} & \small 55.50\textsuperscript{\phantom{(1)}} & \small 66.21 \\
Self-Consistency(GPT3.5)\cite{wang2023selfconsistency} & \small 45.30\textsuperscript{\phantom{(1)}} & \small 90.80\textsuperscript{\phantom{(1)}} & \small 41.80\textsuperscript{\phantom{(1)}} & \small 45.40\textsuperscript{\phantom{(1)}} & \small 55.83 \\
COT(GPT3.5)\cite{sun2023thinkongraph} & \small 42.90\textsuperscript{\phantom{(1)}} & \small 90.10\textsuperscript{\phantom{(1)}} & \small 32.00\textsuperscript{\phantom{(1)}} & \small 28.80\textsuperscript{\phantom{(1)}} & \small 48.45
 \\
\midrule
\multicolumn{6}{c}{w.t. Knowledge~/~Fine-tuned} \\
\midrule
SOTA & \small 45.40\textsuperscript{1}\text{ } & \small 88.20\textsuperscript{2}\text{ } & \small \textbf{87.70}\textsuperscript{3}\text{ } & \small 44.74\textsuperscript{4}\text{ } & \small 66.51
 \\
\midrule
\multicolumn{6}{c}{w.t. Knowledge~/~Zero-Shot~(GPT-4)} \\
\midrule
TOG-R~\cite{sun2023thinkongraph} & \small 54.70\textsuperscript{\phantom{(1)}} & \small 95.40\textsuperscript{\phantom{(1)}} & \small 75.50\textsuperscript{\phantom{(1)}} & \small 86.90\textsuperscript{\phantom{(1)}} & \small 78.13 \\
TOG~\cite{sun2023thinkongraph} & \small 53.80\textsuperscript{\phantom{(1)}} & \small \textbf{95.60}\textsuperscript{\phantom{(1)}} & \small 77.10\textsuperscript{\phantom{(1)}} & \small 88.30\textsuperscript{\phantom{(1)}} & \small 78.70 \\
\kas (Ours) & \small \textbf{66.67}\textsuperscript{\phantom{(1)}} & \small 95.19\textsuperscript{\phantom{(1)}} & \small 86.00\textsuperscript{\phantom{(1)}} & \small \textbf{95.30}\textsuperscript{\phantom{(1)}} & \small \textbf{85.79} \\ 
\bottomrule
\end{tabularx}
\caption{Performance Comparison of different methods. Bold scores stand for best performances among all GPT-based zero-shot methods. The fine-tuned SOTA includes: 1: SPARQL-QA\cite{borroto2022sparqlqa}, 2:  RACo\cite{yu-etal-2022-retrieval}, 3: Re2G\cite{glass-etal-2022-re2g}, 4:RAG\cite{petroni-etal-2021-kilt}.}
\label{tab:main}
\end{table*}
% \vspace{-10pt}

\section{Discussion}

To better understand the key factors influencing our \ka, we conducted extensive analysis experiments. To conserve computational resources, we kept the previously mentioned datasets (\texttt{QALD10-en}, \texttt{Creak}, \texttt{T-REx}, and \texttt{Zero-Shot RE}) but randomly sampled 400 examples each for \texttt{Creak}, \texttt{T-REx}, and \texttt{Zero-Shot RE}.

\subsection{Effect of Observation}

To assess the efficacy of the observation module, we conducted comprehensive experiments with the model without observation. During the action step, \kas selects the action only based on the memory. Subsequently, the reflection step reflects on the triples outputted by the action and updates memory without the guide from observation.

A statistical comparison was performed to evaluate the performance of the \kas with and without observation across all datasets (see Table \ref{tab:ablation}). The results show that the \kas with observation outperforms the \kas without observation, with an average improvement of 3.14\%. Specifically, for \texttt{QALD10-en}  dataset, the \kas with observation outperforms the \kas without observation by 5.41\%. Since \texttt{QALD10-en}  involves multi-hop reasoning, the improved performance of the \kas with observation indicates that the observation module enhances the reasoning ability of the agent, enabling more accurate action selection and reflection.

We can further illustrate the benefits of the observation module with a practical case. In this scenario (see Table \ref{tab:case2}), question is \textit{Where are both The Call of the Wild and White Fang set, the most two famous works of Jack London?}. Without observation, \kas generated the memory, such as \textit{(The Call of the Wild, narrative location, Canada)}, ultimately produced the wrong answer of \textit{Canada}. However, with the observation module, the \kas correctly reasons the memory,such as \textit{(The Call of the Wild, narrative location, Yukon), (White Fang, narrative location, Yukon)}. As a result, the \kas with observation provides the correct answer, \textit{Yukon}. This case exemplifies how the observation module improves the accuracy of action selection and reflection, consequently enhancing the reasoning ability of \kas.

% We can further illustrate the benefits of the observation module with a practical case. In this scenario (see Table \ref{tab:case1}), question is \textit{Where are both The Call of the Wild and White Fang set, the most two famous works of Jack London?}. Without observation, \kas generated the memory, such as \textit{(The Call of the Wild, narrative location, Canada),(Canada, ethnic group, French Canadians)}, ultimately produced the wrong answer of \textit{Canada}. However, with the observation module, the \kas correctly reasons the memory,such as \textit{(The Call of the Wild, narrative location, Yukon),(The Call of the Wild,author,Jack London),(Jack London,notable work,White Fang), (White Fang, narrative location, Yukon)}. As a result, the \kas with observation provides the correct answer, \textit{Yukon}. This case exemplifies how the observation module improves the accuracy of action selection and reflection, consequently enhancing the reasoning ability of \kas.

By incorporating observation information, \kas reasoning power undergoes a dramatic leap, therefore generate an accurate answers. This boost stems from the synergistic interplay between the observation module, harnessing the KG's autonomous reasoning capabilities, and LLM, which further amplifies those strengths. 

\begin{table*}[htbp]
\centering
\small
% \setstretch{1} % 调整行间距
\begin{tabularx}{\textwidth}{l*{7}{>{\centering\arraybackslash}X}}
\toprule
Method & \texttt{QALD10-en} & \texttt{Creak} & \texttt{T-REx} & \texttt{Zero-Shot RE} & Average \\
\midrule
Without Observation & \small 61.26 & \small 95.50 & \small 82.00 & \small 91.75 & \small 82.63 \\
Similarity-based Reflection & \small 61.26 & \small 95.20 & \small 83.20 & \small 93.50 & \small 83.29 \\
Random-based Reflection & \small 58.56 & \small 89.00 & \small 79.50 & \small 92.00 & \small 79.77 \\
Generated-fact Reflection & \small 63.66 & \small 91.00 & \small 80.00 & \small 93.75 & \small 82.10 \\
\kas & \small \textbf{66.67} & \small \textbf{96.00} & \small \textbf{85.40} & \small \textbf{95.00} & \small \textbf{85.77} \\
\bottomrule
\end{tabularx}
\caption{Ablation Comparison}
\label{tab:ablation}
\end{table*}
% \vspace{-10pt}

\subsection{Effect of Observation on Reflection}

In this section, we discuss the impact of observation on reflection module. Three non-observation reflection methods were designed to verify whether observation can enhance the effectiveness of reflection. The similarity-based involves reflecting on the triples from action steps by calculating similarity. In this approach, triples are first sorted based on the similarity score between the $r+t$ and the question. The top-$K$ triples are then selected and stored in memory for the next iteration. The random-based method randomly picks $K$ triples from the action's output and stores them in memory. Finally, the generated-fact method creates $K$ natural language question-related facts for storage. All methods use a setting of $K = 15$.

Table \ref{tab:ablation} showcases our \kas dominance over all three non-observation methods. It achieved an average accuracy increase of 2.48\% compared to the similarity-based method, 6.00\% compared to the random-based method, and 3.66\% compared to the generated-fact method.

In specific scenarios (see Table \ref{tab:case1}), when answering the question \textit{What is the capital of the prefecture Tokyo?}, the generated-fact method resulted in problematic facts, such as \textit{Tokyo is the capital of Tokyo}, and \textit{Tokyo is the capital of Japan}. These were essentially hallucinations created by the LLM based on the given question, which misled the agent and resulted in incorrect answers. In contrast, the reflection of our \kas leveraging observation yielded factual knowledge, \textit{(Tokyo, instance of, prefecture of Japan)}, \textit{(Tokyo, capital,Japan)} and \textit{(Tokyo, capital, Shinjuku)}, consequently enabling the \kas to answer the question correctly. 

% Similarly, the reflection of similarity-based method omitted the crucial triple \textit{(Höxter, inception, +1975-00-00T00:00:00Z)} and included irrelevant information, hindering the agent's ability to generate correct responses.

The findings of Table \ref{tab:ablation} reveal that observation enables reflection module to generate more accurate memories, which translates to improved question-answering accuracy for \ka. This result underscores the value of both leveraging KG autonomous reasoning capabilities and fostering deep collaboration between KG and LLMs. 

% A further analysis was conducted to compare the three non-observation reflection methods. Table \ref{tab:ablation} shows that the random-based method consistently scored the lowest, particularly 5.1\% lower than the third, generated-fact approach. The substantial noise introduced by the random-based method appears to impede the agent's ability to answer questions accurately.

\subsection{Performance across Different Backbone Models}
To evaluate the effectiveness of \kas across various backbones, we analyzed its impact on performance in \texttt{T-REx}  and \texttt{QALD10-en}  datasets. We employed three backbones: GPT-3.5, GPT-4 and DeepSeek-V2 \cite{deepseekai2024deepseek}. DeepSeek-V2 stands out as a powerful, economical, and efficient mixture-of-experts language model. Notably, DeepSeek-V2 surpasses the performance of LLaMA3 70B Instruct on standard benchmarks. 

As evidenced by the Table \ref{tab:backbone}, our \kas approach significantly outperformed the direct answering methods using GPT-3.5, GPT-4 and DeepSeek-V2. Notably, \kas demonstrated a remarkable 30.4\% improvement in direct answering performance when utilizing GPT-3.5 model on \texttt{QALD10-en} dataset. This experiment suggests the generalizability of \kas across different LLMs.

\begin{table}[htbp]
\small
\begin{tabular}{p{4.2cm}cc}
\toprule
Method & \texttt{T-REx} & \texttt{QALD10-en} \\
\midrule
Direct answering(GPT3.5) & 37.60 & 44.74 \\
\kas (GPT3.5) & \textbf{68.00} & \textbf{49.71} \\
\midrule
Direct answering(GPT4) & 57.44 & 57.10 \\
\kas (GPT4) & \textbf{86.00} & \textbf{66.67} \\
\midrule
Direct answering(DeepSeek-V2) &32.86  & 41.14 \\
\kas (DeepSeek-V2) & \textbf{62.67} & \textbf{57.36} \\
\bottomrule
\end{tabular}
\caption{Performance comparison using different backbone models}
\label{tab:backbone}
\end{table}
\vspace{-10pt}

% \footnote{\href{https://github.com/simple-wikidata-db}{simple-wikidata-db}}

\section{Related Works}
\paragraph{KG-enhanced LLM}

Knowledge Graph-enhanced Language Models utilize two primary methodologies when tackling tasks that require integration with KGs. The first involves the extraction of relevant triples from KGs in response to posed questions. \citet{wang2023boosting} prompt LLMs to generate explicit knowledge evidence structured as triples, while \citet{jiang2023structgpt} develop specialized interfaces for gathering pertinent evidence from structured data, enabling LLMs to focus on reasoning tasks based on this information. \citet{baek2023knowledge} retrieve facts related to the input question by assessing semantic similarities between the question and associated facts, then prepending these facts to the input. Meanwhile, \citet{li2023chain} iteratively refine reasoning rationales by adapting knowledge from the KG. \citet{wang2023keqing} dissect complex questions using predefined templates, retrieve entities from the KG, and generate answers accordingly. \citet{luo2024reasoning} employs a planning-retrieval-reasoning framework to generate relation paths grounded by KGs, thereby enhancing the reasoning capabilities of LLMs. Recently, graph retrieve augmented generation~(GRAG) has been introduced to retrieve proper knowledge subgraphs rather than triplets~\cite{he2024g, hu2024grag, mavromatis2024gnn}. In GRAG approaches, subgraphs are first encoded into graph embeddings, and then retrieved based on their similarity to the query. \citet{hu2024grag} proposes an additional step of filtering irrelevant entities within each retrieved subgraph. 

The second approach employs an explore-exploit strategy that guides the knowledge utilization process within the graph. \citet{sun2023think} perform an iterative beam search on the KG to identify the most promising reasoning pathways and report the outcomes. \citet{guo2023knowledgenavigator} selectively accumulate supporting information from the KG through an iterative process that incorporates insights from the LLM to address the question. HyKGE~\cite{jianghykge} first generates a hypothesis of the question by an LLM. Then it retrieves knowledge from the KG according to the entities of the hypothesis and answers the question based on the knowledge. To deal with incomplete KG, \citet{xu2024generate} add a generation operation if some knowledge is missing. 

Although these methods utilize the structural knowledge of KG, the searching or retrieving process is driven by the rationale of LLMs to the target question. Our method is the first to incorporate existing patterns in KG as a way to deeply bind the reasoning abilities of both LLM and KG via our novel observation mechanism.

\paragraph{AI Agent}
In the domain of AI agents, \citet{yao2022react} utilize LLMs to interleave the generation of reasoning traces with task-specific actions. \citet{wu2023autogen} propose an adaptable and conversational agent framework. This framework can operate in various modes, leveraging combinations of LLMs, human input, and auxiliary tools, resulting in a flexible and versatile system. \citet{chen2023autoagents} focus on creating expert agents capable of solving complex tasks.

\section{Conclusion}
In this work, we design \kas framework for KG-centric tasks. In \ka, we introduce KG observation mechanism to autonomously combine the reasoning abilities of KG with LLM. We first propose the observation method to mitigate the problem of explosive number of triples in KG when tackling complex tasks. Then we fuse the observation into the action and reflection modules to further enhance the overall performance. We conduct extensive experiments, and the results clearly illustrate the effectiveness of our framework, highlighting its capability to enhance performance across four KBQA datasets, particularly in handling complicated questions.

\section*{Limitation}
Given the diverse nature of KG-related tasks spanning multiple domains and requiring a broad range of capabilities, the implementation of a multi-agent system is promising to significantly enhance task performance. We leave the integration of our observation mechanism with multi-agent system to future work. Additionally, our current method encounters computational challenges when applied to knowledge graphs with missing relations or entities, such as Freebase. Therefore, in our subsequent work, we plan to address these issues and explore the use of Freebase more comprehensively.

% Considering the KG-related tasks cover questions of various domains and require a wide spectrum of abilities, the multi-agent system would solve the task better. We leave the integration of our observation mechanism with multi-agent sysmtem to future work.

\bibliography{main}

\appendix
\clearpage
\section{Case Study}\label{sec:case}
To demonstrate \kas effectiveness, we highlight several representative cases from the \texttt{QALD10-en} dataset. For improved readability, we replaced entity IDs in \MakeUppercase outputs with their corresponding names. We selected two examples to illustrate how \kas utilizes observation to correctly address the question in Table \ref{tab:case1} and Table \ref{tab:case2}.

\begin{table}[htbp]
\centering
\begin{minipage}{\textwidth}
\begin{tabular}{|p{4cm}|p{11cm}|}
\hline
\textbf{Question} & \textit{The question is What is the capital of the prefecture Tokyo ?} \\
\hline
\textbf{Action} & \textbf{Thought}: The question is asking for the capital of the prefecture Tokyo. The candidate entity ID 'Q1490' corresponds to Tokyo. I can see from the observation that there is a triple (Tokyo, capital, Shinjuku) which might answer the question. However, to confirm this, I will execute a GetNeighbor action on 'Q1490' to get all the triples where Tokyo is the head. \\
& \textbf{Action}: GetNeighbor \\
& \textbf{Entity\_id}: \textit{Tokyo} \\
\hline
\textbf{Reflection} & \textbf{Thought}: The question is asking for the capital of Tokyo. From the observation, we can see that Tokyo is the capital of Japan and it is a prefecture of Japan. The capital of Tokyo is Shinjuku. Therefore, we should select the triples that contain this information. \\
& \textbf{Triples}: \textit{(Tokyo, instance of, prefecture of Japan), (Tokyo, capital
, Shinjuku), (Tokyo, capital, Japan)} \\
\hline
\textbf{Action} & \textbf{Thought}:The question is asking for the capital of the prefecture Tokyo. From the reference memory, it is stated that the capital of Tokyo is Shinjuku. Therefore, the answer to the question is Shinjuku. \\
& \textbf{Action}: Answer \\
\hline
Answer& \textit{Shinjuku} \\
\hline
\end{tabular}
\captionsetup{justification=centering}
\caption{Case 1}
\label{tab:case1}
\end{minipage}
\end{table}
\clearpage
\begin{table}[htbp]
\centering
\begin{minipage}{\textwidth}
\begin{tabular}{|p{4cm}|p{11cm}|}
\hline
Question & \textit{Where are both The Call of the Wild and White Fang set, the most two famous works of Jack London?} \\
\hline
Action & \textbf{Thought}: The question asks about the setting of two works, The Call of the Wild and White Fang, both by Jack London. The observation provides some information about the narrative locations of these works, but to confirm and provide a specific answer, I will use the GetNeighbor function on The Call of the Wild. \\
& \textbf{Action}: GetNeighbor \\
& \textbf{Entity\_id}: \textit{The Call of the Wild} \\
\hline
Reflection & \textbf{Thought}: First, I will focus on the triples related to the locations of the two works, The Call of the Wild and White Fang, as the question is asking about the settings of these two novels. I will also consider the triples related to the author, Jack London, as the question mentions him. I will select the triples that are most directly related to the settings of the two novels. \\
& \textbf{Triples}: \textit{(The Call of the Wild, Narrative location, Yukon), (The Call of the Wild, Narrative location, Canada), (The Call of the Wild, author, Jack London)} \\
\hline
Action & \textbf{Thought}:The question asks for the setting of both The Call of the Wild and White Fang, two notable works of Jack London. From the observation, we can see that The Call of the Wild has narrative locations in the United States of America, Alaska, Canada, and Yukon. However, we don't have the narrative location for White Fang. Therefore, I will use the GetNeighbor function on the entityID of White Fang to find its narrative location.\\
& \textbf{Action}: GetNeighbor \\
& \textbf{Entity\_id}: \textit{White Fang} \\
\hline
Reflection & \textbf{Thought}:First, I will look for triples that directly answer the question. The question is asking for the setting of two works by Jack London, namely, The Call of the Wild and White Fang. Therefore, I will look for triples that contain these two works and a relation that indicates a setting or location. From the candidate triples, I can see that there are two triples that indicate a narrative location for White Fang: (White Fang, Narrative location, Yukon), (White Fang, Narrative location, Canada).\\
& \textbf{Triples}: \textit{(White Fang, Narrative location, Yukon), (White Fang, Narrative location, Canada)} \\
\hline
Action & \textbf{Thought}:The question is asking for the locations where both The Call of the Wild and White Fang are set. From the reference memory, I can see that The Call of the Wild is set in the United States of America, Alaska, Canada, and Yukon. White Fang is set in Canada and Yukon. The common locations for both books are Canada and Yukon.\\
& \textbf{Action}: Answer \\
\hline
Answer& [\textit{Canada},\textit{Yukon}] \\
\hline
\end{tabular}
\captionsetup{justification=centering}
\caption{Case 2}
\label{tab:case2}
\end{minipage}
\end{table}

\clearpage
\section{Prompt}\label{sec:prompt}
This section presents the prompts for the action and reflection modules in tables \ref{tab:actionprompt} and \ref{tab:reflprompt}.
\begin{table}[ht]
\centering
\begin{minipage}{\textwidth}
\begin{tabular}{p{4cm}|p{11cm}}
\hline
\textbf{Action} & \textbf{Prompt} \\ 
\hline
\multirow{3}{4cm}{Neighbor Exploration \& Path Discovery} &Agent Instructions:\\
&You function as an agent that provides answers based on a knowledge graph. \\
&To assist you in querying the KB, use the following tools:\\
& GetNeighbor(entityID: str) -> List[Tuple[str, str, str]]: \\& 
Description: Returns triplets containing the given entityID as the head and its corresponding entityID as the tail. \\&
GetPath(entityID1: str, entityID2: str) -> List[List[Tuple[str, str, str]]]:\\&
Description: Returns all triplets linking the two given entityIDs.\\&
Example Usage:GetPath("Q30", "Q25231") returns all triplets connecting 'Q30' and 'Q25231'.\\
&Data Provided to You:\\&
Question:[Question]\\&
Memory: [Memory]\\&
Candidate EntityIDs: [Task-relevant EntityIDs] (Choose 1 or 2 based on the action)\\&
Observation: [Observation] (These serve as a reference to assist you in selecting the appropriate entityID from the Candidate EntityIDs)\\&
Labels: [Task-relevant entities labels]\\&
Action History: [historical action] (Avoid these actions)\\&
Guidelines:\\&
Choose only one action at a time.\\&
For GetPath, select two entityIDs. For GetNeighbor, select one entityID.
If there are less than 2 entityIDs available, only choose the GetNeighbor action.\\
\hline
Answer & You are a agent that answer questions based on the reference memory and your knowledge. \\
& Here are the reference memory:[Memory]. You can use it to help you answer the quesiton.\\
&Here is the question you are asked to answer the question:[Question].\\
&Ensure that your answer contains one answer or a list of answer, and each answer should be only one or several words,a phrase, a number,true or false, or a date, no other information or descripation in answer.\\
\hline
\end{tabular}
\captionsetup{justification=centering}
\caption{Action Prompt Description}
\label{tab:actionprompt}
\end{minipage}
\end{table}

\begin{table*}[tbp]
\centering
\begin{minipage}{\textwidth}
\begin{tabular}{{p{4cm}|p{11cm}}}
\hline
\textbf{Field} & \textbf{Prompt} \\ \hline
Reflection         & You are an agent that provides answers based on a KG. \\
               & You queried some candidate triples [triples] from last action step and their corresponding labels:[entities labels] from the KB based on the question: [Question]. \\
               & Now you are asked to select related triples, so you can answer the question in the future by using them. \\&
               Here are the observation: [Obervation] for guiding you to select the right triples from the candidate triples. \\&
                Also, here is the memory: [Memory]. You can use it to help you select the right triples from the candidate triples. \\&
                Guidelines:\\&
                You can select less than 15 triples from the candidate triples.\\&
                Your output triples must be in the format of entityID,relationID,entityID.\\
                
\hline
\end{tabular}
\caption{Reflection Prompt Description}
\label{tab:reflprompt}
\end{minipage}

\end{table*}

\end{document}